\documentclass{article}
\usepackage[preprint]{neurips_2024}
\author{
Chen Yang\textsuperscript{1},\;
Ran Le\textsuperscript{1},\;
Yun Xing\textsuperscript{1},\;
Zhenwei An\textsuperscript{1},\;
Zongchao Chen\textsuperscript{1},\\
\textbf{
Wayne Xin Zhao\textsuperscript{2},\;
Yang Song\textsuperscript{1}\thanks{Corresponding author.},\;
Tao Zhang\textsuperscript{1}\;
}
\\
\textsuperscript{1}Nanbeige Lab, BOSS Zhipin\;\\
\textsuperscript{2}Gaoling School of Artificial Intelligence, Renmin University of China\;
\\
{\{yangchen02, leran, songyang\}@kanzhun.com,\; batmanfly@gmail.com}
}

\usepackage{neurips_2024}

\usepackage[utf8]{inputenc} % allow utf-8 input
\usepackage[T1]{fontenc}    % use 8-bit T1 fonts
\usepackage{hyperref}       % hyperlinks
\usepackage{url}            % simple URL typesetting
\usepackage{booktabs}       % professional-quality tables
\usepackage{amsfonts}       % blackboard math symbols
\usepackage{nicefrac}       % compact symbols for 1/2, etc.
\usepackage{microtype}      % microtypography
\usepackage{xcolor}         % colors
\usepackage{minted}
\usepackage{tcolorbox}
\tcbuselibrary{listings, skins}
\tcbuselibrary{breakable}
\usepackage{algorithm}
\usepackage{algpseudocode}
\usepackage{amsmath}
\usepackage{url}
\usepackage{graphicx}
\usepackage{amsmath}
\usepackage{amsthm}
\usepackage{booktabs}
\usepackage{color}
\usepackage{xcolor}
\usepackage{xspace}
\usepackage[misc]{ifsym}

\usepackage[ruled, vlined, nofillcomment, linesnumbered, algo2e]{algorithm2e}
\usepackage{booktabs}
\usepackage{multirow}
\usepackage{balance}
\usepackage{enumitem}
\usepackage{stfloats}
\usepackage{diagbox}
\usepackage{fancyhdr}
\usepackage{stfloats}
\definecolor{result_color}{RGB}{250,250,210}
\usepackage{bm}

\newcommand{\eg}{\emph{e.g.,}\xspace}

\newcommand{\ie}{\emph{i.e.,}\xspace}
\newcommand{\ignore}[1]{}
\newcommand{\paratitle}[1]
{\vspace{1.5ex}\noindent\textbf{#1}}

\title{ToolMind Technical Report: A Large-Scale, Reasoning-Enhanced Tool-Use Dataset}

\begin{document}
\noindent\includegraphics[height=0.8cm]{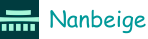}
% \vspace{1cm}

\maketitle

\begin{abstract}
Large Language Model~(LLM) agents have developed rapidly in recent years to solve complex real-world problems using external tools.
However, the scarcity of high-quality trajectories still hinders the development of stronger LLM agents.
Most existing works on multi-turn dialogue synthesis validate correctness only at the trajectory level, which may overlook turn-level errors that can propagate during training and degrade model performance.
To address these limitations, we introduce ToolMind, a large-scale, high-quality tool-agentic dataset with 160k synthetic data instances generated using over 20k tools and 200k augmented open-source data instances.
Our data synthesis pipeline first constructs a function graph based on parameter correlations and then uses a multi-agent framework to simulate realistic user–assistant–tool interactions.
Beyond trajectory-level validation, we employ fine-grained turn-level filtering to remove erroneous or suboptimal steps, ensuring that only high-quality reasoning traces are retained.
This approach mitigates error amplification during training while preserving self-corrective reasoning signals essential for robust tool-use learning.
Models fine-tuned on ToolMind show significant improvements over baselines on $\tau$-bench, $\tau^2$-bench, and BFCL-v4. The dataset is available at \url{https://huggingface.co/datasets/Nanbeige/ToolMind}.

\end{abstract}

\begin{figure}[H]
	\centering
	\includegraphics[width=1\textwidth]{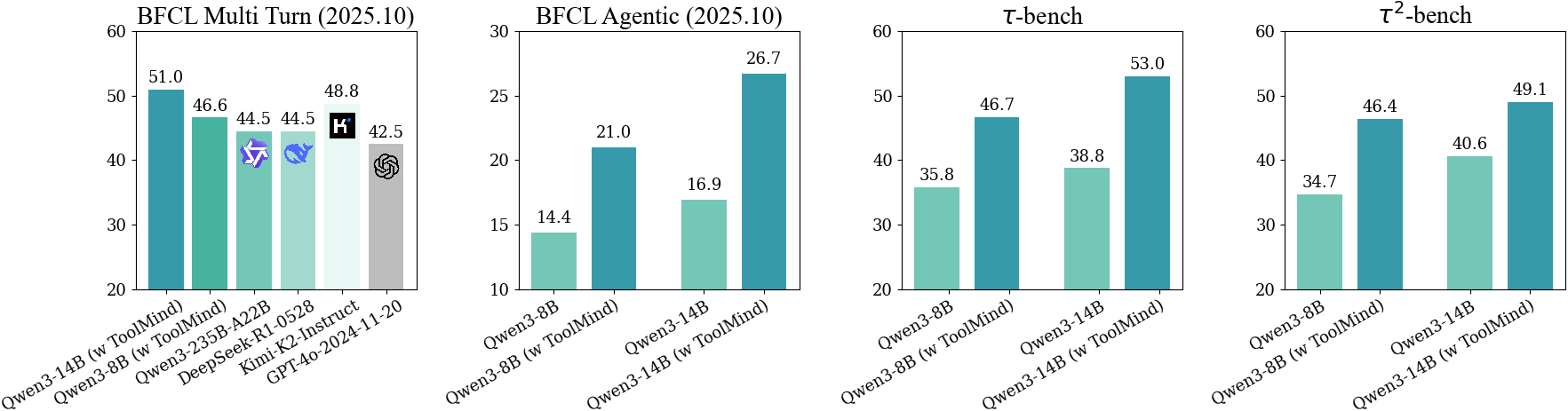}
	\caption{\textcolor{black}{Performance on BFCL-v4, $\tau$-bench, and $\tau^2$-bench.
 }}
	\label{fig:toolmind_performance}
\end{figure}

\section{Introduction}\label{sec:intro}

Large Language Models~(LLMs) have adopted Function Calling~(FC) paradigm to interact with external environments, significantly expanding their capacity to solve complex real-world tasks~\cite{zhao2023survey, qu2025tool}.
% This capability fundamentally relies on the model’s proficiency in tool use, \ie its ability to understand and employ tools robustly, which places higher demands on reliably interpreting user instructions, selecting appropriate functions, and generating precise parameterizations.
This capability hinges on the model’s proficiency in tool use, which requires reliable interpretation of user instructions, appropriate function selection, and precise parameter generation.
A data-centric approach has proven effective in enhancing such capabilities, typically requiring large-scale, high-quality datasets with diverse and accurate function-call demonstrations.

Recent studies have actively synthesized data for various function calling scenarios~\cite{liu_toolace_2024, xi_agentgym_2024}, spanning diverse application domains and interaction modalities such as selection from large toolsets, parallel function calls, and rejection of unspecified requests~\cite{liu_apigen_nodate, prabhakar_apigen-mt_2025, xu_toucan_2025}.
However, existing datasets have several critical limitations: limited scale, the absence of explicit reasoning traces, a lack of multi-turn interactions, and insufficient query diversity. These issues highlights the importance of high-quality, large-scale datasets to develop robust and generalizable tool-use models and agents.

A key challenge in constructing high-quality datasets lies in capturing the dynamic nature of real-world user–assistant interactions.
First, user requests are often under-specified, lacking sufficient parameters for task completion~(e.g., a query like ``What's the weather in Beijing?'' does not specify a time frame). 
Second, effective interaction requires an ongoing clarification process, in which the assistant can proactively request missing information from the user. 
Current open-source synthetic datasets rarely capture these dynamics, leading to models struggling with such scenarios.

In this work, we introduce ToolMind, a reasoning-enhanced tool-use dataset comprising 360k samples. It integrates 160k assistant response turns synthesized through multi-agent simulations and 200k augmented turns adapted from open-source data.  
To ensure diversity, we collect 20k functions from open-source datasets and construct a function graph that captures inter-parameter correlations.
Function chains are then sampled via random walks on this graph to synthesize user intents.
Then, we employ a multi-agent framework for dialogue generation, orchestrating three dedicated roles—user, agent, and tool—each instantiated by a language model to capture the dynamics and complexities of real-world tool-use interactions.
Finally, we apply a rigorous two-stage filtering process, comprising both trajectory-level and turn-level quality checks, to ensure that all training samples contain only valid conversation turns.

Our experiments demonstrate the effectiveness of the ToolMind dataset in enhancing the tool-use capabilities of LLMs. With supervised fine-tuning~(SFT), models of various scales show significant improvements over baselines across multiple benchmarks, including BFCL-v4~\cite{patil_berkeley_nodate}, $\tau$-bench~\cite{yao_-bench_2024}, and $\tau^2$-bench~\cite{barres2025tau}. Our main contributions are summarized as follows:

\begin{itemize}[leftmargin=10pt]
    \item \textbf{High-Quality Tool-Use Dataset.} We introduce ToolMind, a large-scale, high-quality tool-use dataset comprising 360k samples that effectively addresses the key limitations of existing datasets—namely in scale, reasoning traces, multi-turn interactions, and query diversity.
    \item \textbf{Data Synthesis and Quality Filtering Pipeline.} We develop a novel data synthesis pipeline that constructs a function graph based on parameter correlations and employs multi-agent simulations with three dedicated roles (a user, an assistant, and a tool), which better capture the dynamic nature of real-world tool-use interactions. A rigorous two-stage filtering process is further applied to ensure high data quality.
    \item \textbf{Effectiveness of Large-Scale Simulation.} We show that ToolMind delivers significant and consistent performance improvements across models of various scales. These results empirically validate that large-scale synthetic data generated from simulated environments can serve as an effective alternative to real-world data for achieving substantial gains in model performance.
\end{itemize}

\section{Background}
\label{sec:related_work}

\paratitle{Function Calling.}
The integration of external tools and functions into language model workflows marks a major step toward extending model capabilities beyond text generation.
Function calling refers to the structured mechanism through which language models invoke predefined functions or APIs to access real-time data or interact with external systems \cite{qu2025tool}. 
An early paradigm for using external tools is React~\cite{yao2022react}, where models generate both reasoning traces and corresponding function calls in an interleaved manner.
This paradigm was later refined and integrated into thinking models as a native capability~(\eg supporting function calling in chat templates) for training and inference~\cite{yang2025qwen3, zeng2025glm}, also known as the model's function call pattern.
Our synthetic data aims to enhance the model's native function calling capabilities.

\paratitle{Benchmarks.} 
Several benchmarks have been proposed to evaluate the tool-use ability of language models, \ie their capacity to call appropriate functions for task completion~\cite{liu_agentbench_2023, qin_toolllm_2023, guo_stabletoolbench_2025, patil_berkeley_nodate, yao_-bench_2024, barres2025tau, zhong_complexfuncbench_2025}.
These benchmarks have evolved from basic tool-use evaluations to assessing agent behavior in more realistic, interactive scenarios. 
BFCL-v4~\cite{patil_berkeley_nodate} comprises four distinct versions that systematically progress from single-turn interactions to multi-turn dialogues, and from simple, static tool-use cases to complex tasks involving search and memory capabilities. Concurrently, $\tau$-bench~\cite{yao_-bench_2024} focuses on sustained dialogue and function-calling tasks and $\tau^2$-bench~\cite{barres2025tau} further enriches this scenario by extending function access to the user side. Together, these advances demonstrate a clear shift toward more dynamic and complex evaluation frameworks that better reflect real-world conditions. In this work, we evaluate models across the three benchmarks mentioned above to systematically assess the effectiveness of our synthesized data.
% }

\section{Methodology}
\label{sec:method}

In this section, we present our data synthesis pipeline, including graph construction and function-chain sampling, multi-agent, multi-turn trajectory synthesis, and two-stage quality filtering.
% Specifically, we first enhance all functions by leveraging a powerful language model to supplement both the description and the type of their input and output parameters. Based on this enriched representation, we establish connections between functions through their parameter correlations, thereby constructing a large-scale function relation graph. Subsequently, we perform random walks on this graph to obtain numerous function chains that preserve topological relationships. Based on this function chain, we synthesized user intent and generated multi-turn tool-agent-user trajectories through a multi-agent approach.
The overall synthesis pipeline is illustrated in Figure~\ref{fig:pipeline}. All prompts used for trajectory synthesis and quality filtering are provided in Appendix~\ref{app:prompt}.

\begin{figure*}[t!]
	\centering
	\includegraphics[width=1\textwidth]{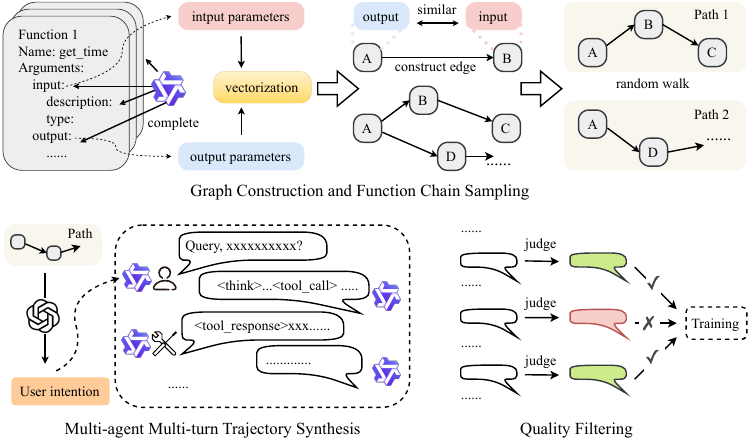}
	\caption{\textcolor{black}{An illustration of the proposed data synthesis pipeline, with three key components: 
    (1)~Graph Construction and Function Chain Sampling, including function collection and refinement, function-graph construction, and random-walk sampling; (2) Multi-Agent Multi-Turn Trajectory Synthesis, using language models to simulate multiple roles and generate interactive trajectories; and (3) Quality Filtering, applying both trajectory-level and turn-level filtering to ensure data quality.
 }}
	\label{fig:pipeline}
\end{figure*}

\subsection{Graph Construction and Function Chain Sampling.}
We construct a directed graph over the collected functions to model their input–output compatibility, and then sample function chains via random walks for trajectory synthesis. 
% The subsections below describe (i) function collection and preprocessing and (ii) random-walk sampling on the graph.

\subsubsection{Functions Collection and Preprocessing} 
Existing open-source works have already achieved substantial tool diversity. Building on these resources, we collect functions from public datasets—including \textit{xlam-function-calling-60k}\footnote{\url{https://huggingface.co/datasets/Salesforce/xlam-function-calling-60k}}
, \textit{glaive-function-calling-v2}\footnote{\url{https://huggingface.co/datasets/glaiveai/glaive-function-calling-v2}}
, and \textit{ToolACE}\footnote{\url{https://huggingface.co/datasets/Team-ACE/ToolACE}}. The resulting function set covers a broad range of function types, complexity levels, and application domains, spanning everyday utilities to domain-specific APIs.
We denote the collected function set by $\mathcal{F}=\{f_i\}_{i=1}^N$, where each function $f_i$ has an input space $\mathcal{X}_i$ and an output space $\mathcal{Y}_i$, and may take multiple arguments, written as:
\[
f_i:\; \mathcal{X}_i \to \mathcal{Y}_i,\qquad 
\mathcal{X}_i=\prod_{k=1}^{m_i}{x}_{i,k},\;\;
\mathcal{Y}_i=\prod_{\ell=1}^{n_i}{y}_{i,\ell}.
\]

To make these function sets compatible for downstream processing, we first standardize incomplete or ambiguous functions (\eg those missing explicit definitions in certain input or output parameters).
We use a language model to infer the missing descriptions and types for each parameter $r\in\mathcal{X}_i\cup\mathcal{Y}_i$, denoted as $\mathrm{desc}(r)$ and $\mathrm{type}(r)$.
Once the specifications are complete, we embed each parameter using a unified scheme. For each $r\in\mathcal{X}_i\cup\mathcal{Y}_i$, we concatenate its description and type with literal tags and encode them with an embedding model:
% \textit{Conan-embedding-vl}~\footnote{https://huggingface.co/TencentBAC/Conan-embedding-v1}:
\[
\mathbf{v}(r)
=\phi\big(\text{\texttt{DESC}}\ \Vert\ \mathrm{desc}(r)\ \Vert\ \text{\texttt{TYPE}}\ \Vert\ \mathrm{type}(r)\big)\in\mathbb{R}^d.
\]

\subsubsection{Random Walk with Function Graph}
With the parameter embeddings in place, we proceed to build a function graph that captures relationships between functions. In this graph, each function is a node, and edges are established based on the semantic fit between outputs and inputs. Concretely, let $G=(V,E)$ be a directed graph with $V=\mathcal{F}$. For two distinct functions $f_i$ and $f_j$, we measure how well an output of $f_i$ can serve as an input to $f_j$ by the maximum cosine similarity over their parameter embeddings:
$$
s_{ij}\;=\;\max_{\,y\in\mathcal{Y}_i,\;x\in\mathcal{X}_j}\operatorname{sim}\!\big(\mathbf{v}(y),\,\mathbf{v}(x)\big),
% ,\qquad 
% \operatorname{sim}(\mathbf{u},\mathbf{v})=\frac{\mathbf{u}^{\top}\mathbf{v}}{\lVert\mathbf{u}\rVert_2\,\lVert\mathbf{v}\rVert_2}.
$$
where $\mathrm{sim}$ denotes cosine similarity. An edge $i\!\to\! j$ is created when this similarity exceeds a fixed threshold $\tau>0$, indicating that an output parameter of $f_i$ is semantically similar to an input parameter of $f_j$. To ensure the reliability of these connections, we further employ a language model as a validator and keep only the accepted links.

We also introduce randomness into edge construction. This yields a more diverse connectivity pattern that better reflects the real-world function composition. With the constructed graph, we sample function chains via random walks that traverse existing edges:
\[
\mathcal{W}=(f_0,f_1,\dots,f_L), L\sim\mathrm{Uniform}\{5,\dots,20\},
\]
where $\mathcal{W}$ is a function chain, $f_t$ denotes the function visited at step $t$, and $L$ is the walking length. To avoid oversampling specific nodes, we impose a per-node visit limit during the random walk process. This encourages balanced graph coverage and mitigates bias toward high-degree hubs.

\subsection{Multi-Agent Multi-Turn Trajectory Synthesis}

Based on the sampled function chains, we synthesize user intents to represent realistic user goals. A user intent represents the underlying goal that a user aims to achieve, rather than the exact query they pose. This distinction provides greater flexibility in modeling user behavior, as users often express the same intent in diverse ways.
Since the tools are not executable in our setup, we employ a simulation framework to generate responses. The trajectories are created through a multi-agent simulation that involves three distinct agents:

\begin{itemize}[leftmargin=10pt]
    \item \textbf{User Agent}: Initiates the interaction by asking questions based on synthesized intent and drives the conversation forward.
    
    \item \textbf{Assistant Agent}: Responds to the user's queries, provides guidance, asks for clarification, or calls external functions when needed.
    
    \item \textbf{Tool Agent}: Simulates the execution of a function invoked by the assistant, returning tool responses based on simulated outputs.
\end{itemize}

These three agents operate in an iterative loop: the user agent poses a query, the assistant agent responds, and the tool agent simulates feedback when required. The process continues until a predefined turn limit or stopping condition is reached. This simulation produces a diverse set of interaction trajectories that serve as training data for modeling rich user–assistant–tool dynamics.

Empirical results demonstrate that even without real-world interaction, the large-scale synthetic data generated from these simulations significantly improves model performance. The flexibility and diversity of the simulated conversations provide valuable insights for training models in complex, multi-agent systems.
For a detailed example of the generated data, see Appendix~\ref{app:data}.

% Once the function chains are sampled, we then synthesize user intents that are not constrained by the sequential task completion order of the chain. Here, user intent is defined as the underlying goal a user seeks to achieve, rather than the specific query they use.
% As we are not in a real environment and the function data comes from external collections, we use a model to simulate the responses due to the lack of genuine feedback.
% The trajectory is generated through a multi-agent simulation, where three models simulate a user, an assistant, and a function: the user poses questions according to the synthesized intent, the assistant responds, and the function provides simulated tool responses. The overall multi-agent synthesis procedure follows the steps summarized in Algorithm~\ref{alg:multiagent}.
% Empirical results show that large-scale, synthetically constructed data from virtual environments yields significant performance, despite the lack of real-world interaction. See Appendix~\ref{app:data} for an example of the data.

\subsection{Quality Filtering}
\label{sec:filter}
To ensure that the synthesized interactions provide reliable learning signals, we apply a two-stage quality filtering process: trajectory-level filtering that maintains goal alignment and coherence, followed by turn-level filtering that removes erroneous or misaligned steps. 

\subsubsection{Trajectory-Level Filtering} 
% We begin by generating user intents that specify the target goals for each interaction.
Given the predefined user intents (defined in the previous section), the user agent first judges whether the task is completed or not during the process of multi-turn interaction. Then, a second LLM evaluates goal coverage, coherence, and cross-turn relevance. Only trajectories that both align with the intent and exhibit a coherent flow are retained, while the rest are discarded. This stage produces a high-quality pool of trajectories for finer-grained filtering.

\subsubsection{Turn-Level Filtering}
Since trajectory-level filtering operates at a coarse granularity, it cannot effectively isolate erroneous turns within otherwise coherent trajectories. Trajectories may still contain erroneous or misaligned turns—such as failed intermediate attempts, tool calls based on incorrect assumptions, or role drift in user/tool messages. To further improve data quality, we address these remaining issues by applying a separate LLM-based judge to each turn.
% Besides, trajectories that pass the trajectory-level checks may contain erroneous or misaligned turns—for example, failed intermediate attempts, tool calls based on incorrect assumptions, or role drift in user/tool messages. 
% Because trajectory-level filtering does not separate valid steps from mistakes, relying on it alone risks training the model to reproduce errors rather than the correction logic. To address this, we perform turn-level annotation and filtering on all retained trajectories, removing invalid or role-inconsistent steps and preserving only correct procedural steps. Together, the trajectory-level and turn-level filters ensure that the final training signal targets effective problem solving and precise tool use.
The model then masks invalid or role-inconsistent steps and retains only correct procedural actions, ensuring the final training data targets effective problem solving and precise tool use.

\subsection{Data Statistics}

\begin{figure}[htbp]
    \centering
    \includegraphics[width=1\textwidth]{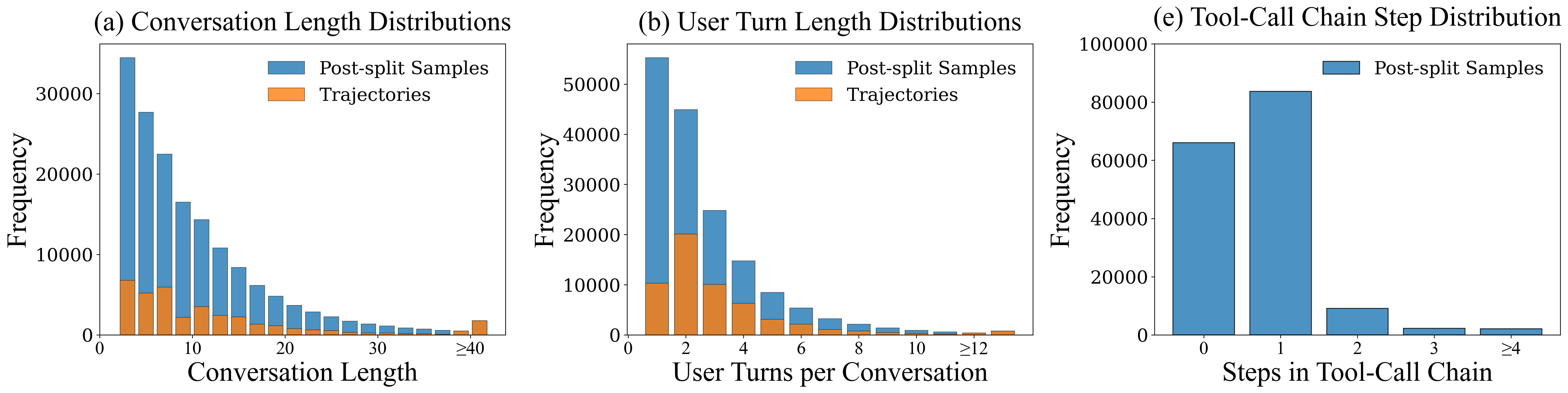}
    \caption{A distribution analysis of the proposed synthetic dataset.}
    \label{fig:dist}
\end{figure}

We split each trajectory into multiple samples using the turns that passed the turn-level quality filter and analyze both \emph{trajectories} (orange) and \emph{post-split samples} (blue).
See the distribution analysis in Figure~\ref{fig:dist}.
From a trajectory-level perspective, the overall distribution of data lengths is relatively uniform, and users show strong participation throughout the dialogues. After the conversations are segmented into turns and filters are applied, the distribution skews toward shorter instances, which is expected. We further analyze the distribution of step counts for consecutive tool calls. A sizable fraction of turns proceed without any tool calls, likely because the synthetic user is prompted to mimic natural dialogue—holding back some constraints so that the model must ask follow-up questions and confirm details. Taken together, these findings indicate that current tasks leave room for greater complexity, motivating future work on more demanding tool-use scenarios.

\begin{figure}[htbp]
    \centering
    \includegraphics[width=0.7\textwidth]{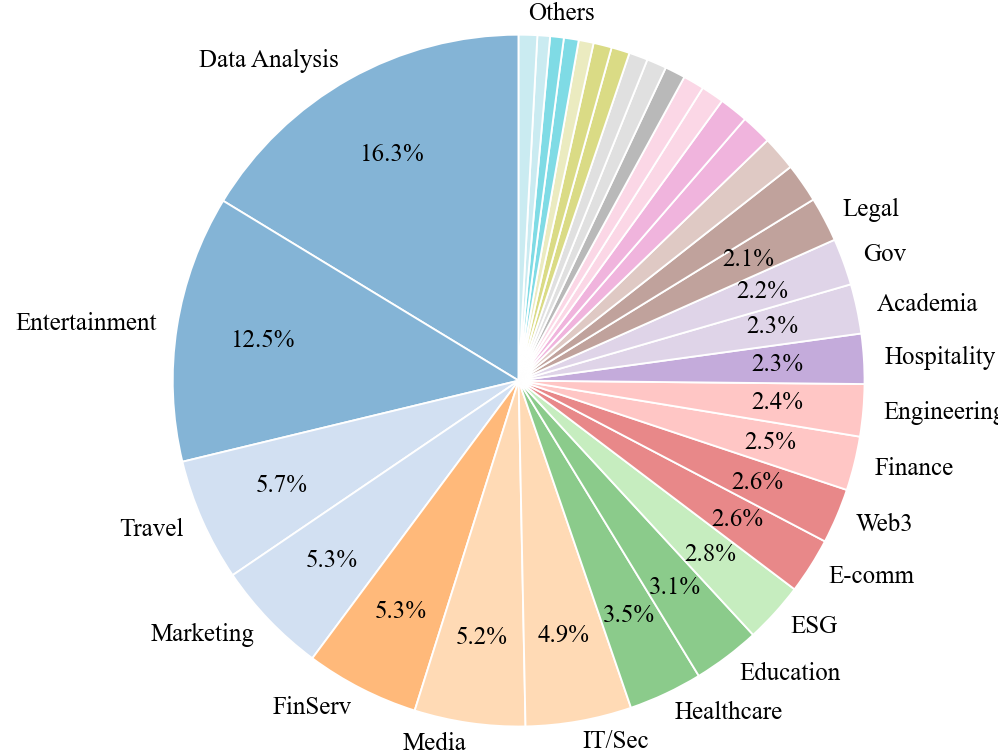}
    \caption{Distribution of user intent domains, with tail ($\le 2\%$) grouped as ``others''.}
    \label{fig:domain}
\end{figure}

In addition, we perform domain classification and statistical analysis of user intents throughout the synthesis process. Because each intent is composed via a function chain, a single intent can have multiple domain labels. As Figure~\ref{fig:domain} shows, the data analysis and entertainment domains are the most prevalent, while the remaining domains are relatively balanced.

% Figure~\ref{fig:dist} shows that conversation length is heavy-tailed (a), with most interactions short and a diminishing tail (bucketed at $\geq\!40$); splitting amplifies counts across all lengths while largely preserving the shape. User turns per conversation display a similar pattern (b): mass concentrates at 1–4 turns with a merged tail at $\geq\!12$, and splitting uniformly increases volume because each valid exchange yields additional samples. For tool use (c), the consecutive tool-call chain step distribution for post-split samples is dominated by 0–1 steps, with multi-step chains rapidly decaying and $\geq\!3$ steps rare. Taken together, these statistics indicate a corpus rich in short tasks yet covering longer dialogues and occasional multi-step compositions; in training, we thus sample broadly while upweighting long conversations and deeper chains to strengthen long-horizon reasoning and tool composition.

\section{Experiments}
\label{sec:experiment}
This section presents the performance of models of varying sizes after supervised fine-tuning~(SFT) on our dataset.
We first present the fine-tuning setup and then evaluate the resulting models against the baselines on widely-used tool-calling benchmarks.

\subsection{Experimental Setup}

\paratitle{Model and Fine-tuning Setup.}
We perform supervised fine-tuning~(SFT) on \texttt{Qwen3-8B}, and \texttt{Qwen3-14B}~\cite{yang2025qwen3}. The training is conducted using OpenRLHF~\cite{hu2024openrlhf} framework with a sequence length of 64k, a global batch size of 64, a learning rate of 5e-6, and a warmup ratio of 0.03. 
% All experiments were performed with NVIDIA A800 GPUs.

% The training data consists of a series of trajectories, each represented as a multi-turn dialogue among the user, the assistant, and tool responses. Each assistant message is composed of three parts: think, content, and tool call. Since the thinking component in each assistant message tends to be lengthy, to mitigate the issue of excessively long multi-turn contexts, we decompose each sample: an n-turn dialogue is split into n single-turn samples, with the thinking segments in the historical context removed. We finally use these split single-round samples for training.

\paratitle{Benchmarks.}
We evaluate our models on three widely used tool-calling benchmarks: BFCL-v4~\cite{patil_berkeley_nodate}, $\tau$-Bench~\cite{yao_-bench_2024}, and $\tau^2$-bench~\cite{barres2025tau}.
For BFCL-v4, we use the official evaluation setup for Qwen3 and benchmark our model against larger-scale models on the official leaderboards as of October 2025. For a comprehensive comparison, we incorporat results from both open-source models \texttt{DeepSeek-V3}~\cite{liu2024deepseek}, \texttt{DeepSeek-R1-0528}~\cite{guo2025deepseek}, \texttt{Qwen3-235-instruct}~\cite{yang2025qwen3}, \texttt{Kimi-K2-Instruct}~\cite{team2025kimi} and closed-source models \texttt{GPT-4o-2024-11-20}, \texttt{GPT5-2025-0807}, and \texttt{Gemini2.5-Pro}~\cite{comanici2025gemini}. For $\tau$-Bench and \mbox{$\tau^2$-Bench}, we used GPT-4o as a user simulator and evaluate performance by directly comparing the model before and after fine-tuning.

\paratitle{Datasets.}
The entire ToolMind dataset comprises two components: (i) synthesized trajectories described in Section~\ref{sec:method} and (ii) data augmented from open-source corpora. % Since the preprocessing process splits the data with the valuable turns, we only calculate the loss for the final assistant response.
We incorporat the following open-source datasets: xlam-function-calling-60k~\cite{liu_apigen_nodate}, When2Call~\cite{ross_when2call_2025}, glaive-function-calling-v2~\footnote{https://huggingface.co/datasets/glaiveai/glaive-function-calling-v2}, ToolACE~\cite{liu_toolace_2024}, BUTTONInstruct~\cite{chen_facilitating_2025}, APIGen-MT-5k~\cite{prabhakar_apigen-mt_2025}, and the Tau-bench training set~\cite{yao_-bench_2024}. 
All open-source multi-turn datasets are subjected to the same split and quality-filtering procedures as the synthesised data.
The statistics of the datasets can be found in Table~\ref{tab:data}.
After preprocessing, each multi-turn conversation is split at each assistant message into multiple samples.
For each splitted sample, we retain the full prior context up to (and including) the anchor assistant message and discard all subsequent turns.
Each asisstant response is structured into three parts: think, content, and tool calls.
During training, we compute loss only on the tokens of the anchored assistant response.

\begin{table}[]
\label{tab:data}
\centering
\caption{Statistical Overview of Data Samples. “Trajectories (Queries)” denote individual interaction sequences collected from datasets, while “Samples after processed” refers to the count of data samples after reconstruction and quality filtering.}
\resizebox{0.8 \textwidth}{!}{
\begin{tabular}{lrrr}
\toprule
Dataset                    & \# Trajectories~(Queries) & \# Samples after processed \\ 
\midrule
xlam-function-calling-60k     &    60,000    &    101,363   \\
When2Call~(sft)            &  15,000  &    17,531  \\
Glaive-function-calling-v2     &  112,960  &    20,017   \\
ToolACE                     &  11,300  &    7,327   \\
BUTTONInstruct                &  8,000  &    21,202  \\
APIGen-MT-5k                &   5,000     &    25,109   \\
Tau-bench train set           &   -   &    12,882   \\ 
\textit{Synthesized trajectories}            &   40,161     &    163,180   \\ 
\midrule
ToolMind                     &   111,941     &    368,611   \\ \bottomrule
\end{tabular}
}
\end{table}

\subsection{Experimental Results}

\begin{table}[]
\centering
\caption{Overall Performance Comparison on BFCL-v4. The term ``FC'' denotes native support for function/tool calling, whereas ``Prompt'' refers to a walk-around implementation that leverages the model's inherent text generation ability. The upper section of the table reports results from the BFCL October leaderboard, while the lower section presents scores evaluated locally.}
% ~\footnotetext{https://gorilla.cs.berkeley.edu/leaderboard.html}
\label{tab:bfcl}
\resizebox{1 \textwidth}{!}{
\begin{tabular}{lcccccc}
\toprule
\multirow{2}{*}{Model} &  \multirow{2}{*}{Overall} & \multicolumn{2}{c}{Single Turn} & \multirow{2}{*}{Multi Turn} &  \multicolumn{2}{c}{Agentic} \\
\cmidrule{3-4} \cmidrule{6-7}
&    & Non-live(AST) & Live(AST) &  & Search & Memory  \\
\midrule
DeepSeek-v3~(FC)        & 45.20 & 88.77 & 79.94 & 33.00  & 32.50 & 22.37  \\
DeepSeek-R1-0528~(FC)   & 48.97 & 75.73 & 80.90 & 44.50  & 63.00 & 0.00    \\
Qwen3-235-instruct~(FC) & 54.37 & 88.10 & \textbf{82.61} & 44.50  & 49.00 & 29.25  \\
Kimi-K2-Instruct~(FC)   & 56.07 & 84.02 & 77.57 & \textbf{48.75} & 59.00 & 25.16\\
GPT-4o-2024-11-20~(FC)  & 50.27 & 83.88 & 70.54 & 42.50  & 40.50 & 28.82\\
GPT5-2025-0807~(FC)     & \textbf{59.22} & 72.92 & 58.25 & 28.50  & \textbf{84.50} & \textbf{57.63}  \\
Gemini2.5-Pro~(Prompt)  & 54.14 & \textbf{89.54} & 76.83 & 30.62 & 66.50 & 31.61  \\
\midrule
\midrule
Qwen3-8b~(FC)        & 42.21 & \textbf{88.27} & 80.83 & 38.88 & 10.00 & 18.71 \\
\quad \textit{with} ToolMind & \textbf{46.92}\textsuperscript{\textcolor{green!60!black}{+4.69\%}} & 88.06 & \textbf{81.42} & \textbf{46.62} & \textbf{21.50} & \textbf{20.43} \\
\midrule
Qwen3-14b~(FC)                & 45.14 & \textbf{90.10} & \textbf{80.90} & 44.12 & 12.50 & \textbf{21.29}  \\
\quad \textit{with} ToolMind & \textbf{50.54}\textsuperscript{\textcolor{green!60!black}{+5.40\%}} & 89.00 & 80.83 & \textbf{51.00} & \textbf{35.50} & 17.85 \\
\bottomrule
\end{tabular}}
\end{table}

\begin{table}[]
\centering
\caption{Overall Performance Comparison on $\tau$-Bench and $\tau^2$-Bench. The table reports average and per-domain scores for different models, with and without ToolMind integration.}
\label{tab:taubench}
\resizebox{0.9 \textwidth}{!}{
\begin{tabular}{lccccccccc}
\toprule
\multirow{2}{*}{Model}&  \multicolumn{3}{c}{$\tau$-bench} & \multicolumn{4}{c}{$\tau^{2}$-bench}  \\
\cmidrule{2-4} \cmidrule{5-8}
&  Avg. & retail & airline & Avg. & retail & airline & telecom \\
\midrule
qwen3-8b~(FC)           & 35.83 & 35.65           & 36.00   &   34.64           & 43.86 & 32.00        & 28.07        \\
\quad \textit{with} ToolMind  &    \textbf{46.70}\textsuperscript{\textcolor{green!60!black}{+10.87\%}}    &   \textbf{57.39}   &    \textbf{36.00}      &  \textbf{46.41}\textsuperscript{\textcolor{green!60!black}{+11.77\%}}     & \textbf{59.65}    &   \textbf{48.00}      & \textbf{31.58}             \\
 \midrule
qwen3-14b~(FC)        &  38.78  &   49.56      &  28.00   &  40.65          & 52.63     & 36.00       & \textbf{33.33}             \\
\quad \textit{with} ToolMind &   \textbf{53.00}\textsuperscript{\textcolor{green!60!black}{+14.22\%}}     &   \textbf{60.00}    & \textbf{46.00}          &   \textbf{49.08}\textsuperscript{\textcolor{green!60!black}{+8.43\%}}    & \textbf{59.65}     &   \textbf{56.00}     & 31.58                 \\
\bottomrule
\end{tabular}}
\end{table}

\paratitle{Overall Performance.} The performance of our SFT models on $\tau$-Bench~\cite{yao_-bench_2024}, and $\tau^2$-bench~\cite{barres2025tau}, BFCL-v4~\cite{patil_berkeley_nodate} is summarized in Table~\ref{tab:bfcl} and Table~\ref{tab:taubench}. 
On BFCL-v4, we observe notable improvements in both Qwen3-8B and Qwen3-14B after training. Although their single-turn evaluation scores show modest fluctuations, they remain above most baseline models. Performance gains are most pronounced in multi-turn and agentic evaluations.
Notably, after training on ToolMind, Qwen3-14B surpasses several larger-scale open-source and closed-source models. Similarly, on the $\tau$-bench and $\tau^2$-bench benchmarks, both Qwen3-8B and Qwen3-14B achieve substantial improvements in their average score after training on ToolMind. 
These results demonstrate the effectiveness of the ToolMind dataset in improving model tool-use capabilities.

\begin{table}[]
\centering
\caption{Ablation Study of ToolMind Components. Models trained on three data configurations are compared: (a) synthesized data only,
(b) synthesized data without turn-level quality filtering, and (c) data augmented from open-source datasets.}
\label{tab:ablation}
\resizebox{1 \textwidth}{!}{
\begin{tabular}{lcccccccc}
\toprule
                         % & Tau1-airline & Tau1-retail & Tau2-airline & Tau2-retail & Tau2-telecom & BFCL-v4 & BFCL-v4-agentic \\
\multirow{2}{*}{Model}&  \multicolumn{3}{c}{$\tau$-bench} & \multicolumn{4}{c}{$\tau^{2}$-bench} & \multicolumn{1}{c}{BFCL-v4} \\
&  Avg.& retail & airline & Avg.& retail & airline & telecom & overall  \\
\midrule
Qwen3-8b~(FC)           & 35.83     &  35.65    &  36.00    &  34.64  &  43.86     &    32.00     & 28.07         & 42.21             \\
\quad \textit{with} (a) synthesized data & 42.31     &  42.61    &  \underline{42.00}    &  38.85  &    42.98   &     42.00    &      \textbf{31.58}    &  \underline{46.87}           \\
\quad \textit{with} (b) no turn-level filtering &  35.31   &  42.61  &  28.00    &  41.73  &   47.37  &   48.00    &    29.82      &     44.11      \\
\quad \textit{with} (c) augmented open-source data &    \textbf{48.65}  &    \underline{51.30}    &  \textbf{46.00}    &  \underline{42.16}  &    \underline{57.89}   &     \underline{44.00}    &        24.56  &   45.88      \\
\quad \textit{with} ToolMind  &    \underline{46.70}    &   \textbf{57.39}   &    36.00      &    \textbf{46.41}  &   \textbf{59.65}     &   \textbf{48.00}      &         \textbf{31.58}     &    \textbf{46.92}       \\
\bottomrule
\end{tabular}}
\end{table}

\paratitle{Ablation Study.} To evaluate the effectiveness of each component in ToolMind, we conduct an ablation study on Qwen3-8B. The study comprises the following configurations: (a) using only synthesized data, (b) using synthesized data without turn-level quality filtering, and (c) using only data augmented from open-source datasets.
The experimental results, shown in Table~\ref{tab:ablation}, demonstrate that the open-source augmented data contribute to significant performance improvements on $\tau$-bench and $\tau^2$-bench, while the graph sampling-based synthetic data lead to major gains on BFCL-v4, and the turn-level quality filtering significantly enhances the model performance. Moreover, the model trained on the combined dataset achieves the best results across nearly all evaluation metrics, demonstrating that each component of ToolMind is integral and effective.

\section{Conclusion}
This paper presents ToolMind, a large-scale dataset designed to address data scarcity in tool-use for Large Language Models. Built through graph-based function sampling and multi-agent trajectory synthesis, ToolMind offers diverse, realistic function-calling trajectories with explicit reasoning traces. Experimental results demonstrate that models fine-tuned on ToolMind achieve consistent and significant improvements across multiple benchmarks. Our work highlights the value of high-quality synthetic data for advancing tool-use capabilities and provides a valuable resource for future research.

\bibliographystyle{plain}  % 引用样式
\bibliography{references}

\newpage
\appendix
\section{Prompts}
\label{app:prompt}
\subsection{Prompt for completing functions}
\label{app:complete_function}
This section details the prompt for completing functions and determining correlations.

\newtcolorbox{promptbox}{
  breakable,
  enhanced,
  colback=blue!10,          % 浅蓝色背景
  colframe=blue!30,         % 可选，边框颜色
  coltext=black,            % 字体黑色
  boxrule=0.5pt,            % 边框粗细
  arc=2mm,                  % 圆角
  left=2mm, right=2mm, top=1mm, bottom=1mm,
}

% 使用示例
\begin{promptbox}
Please help me predict the input of the function. 

\vspace{2mm}  % 2毫米空行

Return only one result in JSON format with two fields: input description and input structure. 
Input description should describe the content of the input, while input structure should be a list of parameters, each with a name, a description and a type.

\vspace{2mm}  % 2毫米空行

The function is: \{function\}
\end{promptbox}

\begin{promptbox}
Please help me predict the output of this function. 

\vspace{2mm}  % 2毫米空行

Return only one result in JSON format with two fields: output description and output structure. 
Output description should describe the content of the output, while output structure should be a list of parameters, each with a name, a description and a type.

\vspace{2mm}  % 2毫米空行

The function is: \{function\}
\end{promptbox}

\begin{promptbox}
Please determine whether the following "source\_function" and "target\_function" can be highly correlated or the output result of "source\_function" is suitable as the input parameter of "target\_function". If the correlation is large or suitable for parameter passing, the specific function is as follows:
\vspace{2mm}  % 2毫米空行

\#\#\# Source\_function:
\vspace{2mm}  % 2毫米空行

\{source\_function\} 
\vspace{2mm}  % 2毫米空行

\#\#\# Target\_function:
\vspace{2mm}  % 2毫米空行

\{target\_function\}
\vspace{2mm}  % 2毫米空行

Please evaluate the strength of the association edges and assign a score from 0 to 9 for the following aspects: field transitivity, coherence of potential user intent paths. The output should be a JSON object with exactly two fields: "Field transitivity", "Potential user intent path coherence". Do not output anything else!!!
\end{promptbox}

\subsection{Prompt for trajectory synthesis}
\label{app:trajectory_synthesis}
This section details the prompts for trajectory synthesis, including prompts for user and tool simulator.

\subsubsection{Prompt for synthesis user intent}

\begin{promptbox}
Suppose you have another assistant who has access to the following tools to get information. Please generate one task instruction that mimic real human users and their intentions, such as having different personalities and goals. Note that the intent should be as natural as possible, covering as many tools as possible, but not forcing overwriting if the tools are not closely related. Please ignore image-related tools and do not generate image-related instructions. User intents should be highly consistent, avoiding the awkward patchwork of several unrelated tasks. 
\vspace{2mm}  % 2毫米空行

\#\#\# Tools
\vspace{2mm}  % 2毫米空行

\{tools\}
\vspace{2mm}  % 2毫米空行

Please output the results strictly in JSON format with two fields: "Task Instruction" and "Tool Usage" and don't output anything else!!
\end{promptbox}

\subsubsection{Prompt for user simulator}
\begin{promptbox}
You are a human user and must act as a genuine user throughout the conversation, interacting in a manner consistent with normal human behavior. 
Your primary goal is to achieve the following intent by seeking guidance, advice, or assistance from other participants.
\vspace{2mm}  % 2毫米空行

\#\#\# Your intent:
\vspace{2mm}  % 2毫米空行

\{intent\}
\vspace{2mm}  % 2毫米空行

Please adhere strictly to the following guidelines:
\vspace{2mm}  % 2毫米空行

1. Role Consistency and Natural Interaction: Always maintain the role of a user. Do not respond as an assistant, AI, or any authoritative figure. Speak naturally, as a real human would. Avoid repetitive, mechanical, or overly structured responses.
\vspace{2mm}  % 2毫米空行

2. Incremental Disclosure: Do not reveal your entire intent at once. Unfold your needs gradually over multiple turns. Use common human conversation strategies, such as showing uncertainty when appropriate.
\vspace{2mm}  % 2毫米空行

3. Response to Fulfillment: If the the other participant successfully fulfills your intent, output '\#\#\#STOP\#\#\#' immediately. Do not output '\#\#\#STOP\#\#\#' in the first turn, regardless of the conversation!!
\vspace{2mm}  % 2毫米空行

\end{promptbox}

\subsubsection{Prompt for tool simulator}

% 使用示例
\begin{promptbox}
You are simulating a high-performance computer system with complete computational capabilities. You have access to extensive external knowledge, can execute any arbitrary function, and operate without errors. For a given function, you should simulate the execution of a computer system program as accurately as possible.
\vspace{2mm}  % 2毫米空行

\#\#\# Function info

\vspace{2mm}  % 2毫米空行
\{function\_info\}
\vspace{2mm}  % 2毫米空行

\#\#\# Function call
\vspace{2mm}  % 2毫米空行

\{tool\_calls\}
\vspace{2mm}  % 2毫米空行

Given this function call, you should execute the function and return the results strictly in JSON format. 
Your response should contain only the JSON result, without any additional or irrelevant text.
The result must be enclosed within <func\_return> and <\/func\_return> tags.
If the function call is invalid (e.g., incorrect function name, missing or malformed arguments), return a JSON error message clearly indicating the cause.
\vspace{2mm}  % 2毫米空行

\#\#\# Example of function call and function return: 
\vspace{2mm}  % 2毫米空行

[{{ "name": "get\_weather", "arguments": {{"city": "New York"}} }}] 
\vspace{2mm}  % 2毫米空行

<func\_return> {{ "temperature": "25$^{\circ}$C" }} <\/func\_return>

\end{promptbox}

\subsection{Prompt for quality filtering}
\label{app:quality_filtering}
This section details the prompts used for quality filtering, including trajectory-level filtering and turn-level filtering.

\subsubsection{Prompt for trajectory-level quality filtering}
\begin{promptbox}
Please strictly evaluate the quality of the following multi-turn dialogue data based on the following criteria: contextual coherence, role consistency, logical soundness, and accuracy of tool usage.
Your task is to make a binary judgment — if the dialogue is of good quality, output 1; otherwise, output 0.
\vspace{2mm}  % 2毫米空行

\#\#\# Tools
\vspace{2mm}  % 2毫米空行

\{tools\}
\vspace{2mm}  % 2毫米空行

\#\#\#Multi-turn conversations
\vspace{2mm}  % 2毫米空行

\{messages\}
\vspace{2mm}  % 2毫米空行

Please make a strict and comprehensive assessment of the dialogue’s quality, considering whether it maintains contextual coherence, consistent role behavior, logical reasoning, and correct use of tools.
Finally, output only a single digit: 0 or 1. Do not include any other text or explanation.
\end{promptbox}

\subsubsection{Prompt for turn-level quality filtering}
\begin{promptbox}
Please strictly evaluate the quality of the last response in the following dialogue data, based on contextual coherence, logical consistency, and accuracy of tool usage.
Determine whether the response is semantically aligned with the previous dialogue, logically sound without contradictions, and employs the mentioned tools correctly according to their definitions and argument structures.
If the response is of good quality, output 1; otherwise, output 0.
\vspace{2mm}  % 2毫米空行

\#\#\# Tools mentioned in the conversation
\vspace{2mm}  % 2毫米空行

\{tools\}
\vspace{2mm}  % 2毫米空行

\#\#\# Conversation history
\vspace{2mm}  % 2毫米空行

\{messages\}
\vspace{2mm}  % 2毫米空行

\#\#\# Last response
\vspace{2mm}  % 2毫米空行

\{response\}
\vspace{2mm}  % 2毫米空行

Please make a strict judgment on whether the last response is of good or poor quality, considering contextual coherence, logical soundness, and correctness of tool usage.
Finally, output only a single digit: 0 or 1. Do not output any other text or explanation.
\end{promptbox}

\section{Examples}
\label{app:data}

\begin{promptbox}
\#\#\# Function Chains
\vspace{2mm}  % 2毫米空行

\{'name': 'getStockLocations',
'description': 'Retrieves a list of stock locations for an eCommerce application.',
'parameters': \{'type': 'dict',
'properties': \{'limit': \{'description': 'Limits the number of items on a page (max 100).',
'type': 'float'\}\},
'required': []\},
'required': None\}
\vspace{2mm}  % 2毫米空行

\{'name': 'checkInventory',
'description': 'Check the inventory of a specific product',
'parameters': \{'type': 'dict',
'properties': \{'product\_code': \{'type': 'string',
'description': 'The code of the product to check inventory for (e.g., ABC123)'\},
'location': \{'type': 'string',
'description': 'The location to check inventory at (e.g., warehouse A, store B)'\}\},
'required': ['product\_code']\},
'required': None\}
\vspace{2mm}  % 2毫米空行

\{'name': 'checkIpAddress',
'description': 'Check if an IP address is safe or not.',
'parameters': \{'type': 'dict',
'properties': \{'ip\_address': \{'type': 'string',
'description': 'The IP address to be checked for safety'\}\},
'required': ['ip\_address']\},
'required': None\}
\vspace{2mm}  % 2毫米空行

\#\#\# User Intent
\vspace{2mm}  % 2毫米空行

I'm a warehouse manager for an eCommerce company. First, I need to check where our inventory is stored by retrieving the stock locations. Then, I want to verify the availability of product 'XYZ789' specifically at 'warehouse C' to plan restocking. Finally, I need to ensure our system's security by checking if the IP address 192.168.1.100 is safe.
\vspace{2mm}  % 2毫米空行

\#\#\# Trajectory
\vspace{2mm}  % 2毫米空行

user: Can you retrieve the current stock locations for me? 

\vspace{2mm}  % 2毫米空行
assistant: <think>Okay, the user is asking to retrieve the current stock locations. Let me check the available tools. There's a function called getStockLocations. The parameters for that include a limit, which is optional and can be up to 100. The user didn't specify a limit, so I don't need to include that. Since the function doesn't require any parameters, I can call it without any arguments. I should generate the tool call with the name getStockLocations and empty arguments. Wait, but the parameters are a dict, so maybe I should pass an empty object. Yeah, that makes sense. So the tool call would be {"name": "getStockLocations", "arguments": {}}. I need to make sure that the JSON is correctly formatted. No other parameters are required, so that's all.</think>
<tool\_call>
\{"name":"getStockLocations", "arguments":"\{\}"\}
</tool\_call>

\vspace{2mm}  % 2毫米空行
tool: \{  "stock\_locations": [    {      "id": 1,      "name": "Warehouse A",      "address": "123 Main St, Cityville",      "capacity": 5000    },    {      "id": 2,      "name": "Store B",      "address": "456 Oak Ave, Townsburg",      "capacity": 2000    },    {      "id": 3,      "name": "Distribution Center C",      "address": "789 Pine Rd, Countryside",      "capacity": 10000    }  ],  "pagination": {    "page": 1,    "total\_pages": 5  }\}

\vspace{2mm}  % 2毫米空行

assistant:
<think>Okay, let me process the user's request. They asked for the current stock locations, so I called the getStockLocations function. The response came back with three locations: Warehouse A, Store B, and Distribution Center C. Each has an ID, name, address, and capacity. The pagination shows there are 5 total pages, but the user might not need all of them right now. I should present this information clearly. Start by listing each location with their details. Maybe use bullet points for readability. Also, mention the pagination part so the user knows there's more data available if needed. They might want to ask for the next page or check a specific product's inventory next. I'll keep the tone friendly and offer further assistance in case they need more details or another action, like checking inventory for a product. Make sure to highlight the key points like names, addresses, and capacities. Alright, let's put that together in a concise response.</think>
Here are the current stock locations retrieved:1. **Warehouse A**     - **Address:** 123 Main St, Cityville     - **Capacity:** 5,000 units  2. **Store B**     - **Address:** 456 Oak Ave, Townsburg     - **Capacity:** 2,000 units  3. **Distribution Center C**     - **Address:** 789 Pine Rd, Countryside     - **Capacity:** 10,000 units  There are **5 total pages** of results (you’re currently viewing page 1). Would you like to:  - Check inventory for a specific product at one of these locations?  - Retrieve the next page of stock locations?

\vspace{2mm}  % 2毫米空行
user: Can you check the availability of product XYZ789 at Distribution Center C?

\vspace{2mm}  % 2毫米空行

assistant: 
<think>Okay, the user is asking to check the availability of product XYZ789 at Distribution Center C. Let me see which tool to use here.Looking at the available tools, there's the checkInventory function. It requires the product\_code and optionally the location. The user specified both the product code and the location, so I need to call checkInventory with those parameters.Product code is XYZ789, and the location is Distribution Center C. Let me make sure the parameters are correctly formatted. The location in the tool might need to be a string, so I'll input "Distribution Center C" as the location. I should structure the tool call with these arguments. Let me double-check if there are any other required parameters. The tool's required field is product\_code, which is provided. Location is optional but the user specified it, so include it.Alright, the tool call should be correct with those parameters.</think>
<tool\_call>
\{"name": "checkInventory",
"arguments": "\{"product\_code": "XYZ789"\}\}
</tool\_call>

\vspace{2mm}  % 2毫米空行
tool: \{"product\_code": "XYZ789",   "location": "Distribution Center C",   "available\_quantity": 150,   "last\_updated": "2023-10-05T14:30:00Z" \}

\vspace{2mm}  % 2毫米空行
assistant: 
<think>Okay, let me process the user's request. They wanted to check the availability of product XYZ789 at Distribution Center C. I called the checkInventory tool with the product code and location. The response shows 150 units available, last updated on October 5th.Now, I need to present this information clearly. The user might be a warehouse manager or someone involved in order fulfillment. They probably need this info to fulfill an order or manage stock levels. I should mention the available quantity and the last updated timestamp to indicate the data's freshness. Also, offer further assistance in case they need more details or actions, like checking another location or placing an order. Keep the response concise but informative.</think>
Here is the inventory availability for product **XYZ789** at **Distribution Center C**:- **Available Quantity:** 150 units - **Last Updated:** October 5, 2023, 2:30 PM UTC  Would you like to check availability at another location or proceed with an order?

\vspace{2mm}  % 2毫米空行
user: Is the IP address 192.168.1.100 flagged as a security risk in our system?

\vspace{2mm}  % 2毫米空行
assistant:
<think>
Okay, the user is asking if the IP address 192.168.1.100 is flagged as a security risk in our system. Let me check the tools available. There's a function called checkIpAddress that takes an IP address as a parameter. I need to call that.First, I'll verify the parameters. The function requires the ip\_address field, which the user provided. The IP given is 192.168.1.100. I should make sure it's a valid IP. Since it's in the correct format, I can proceed. I don't see any other parameters needed for this function. So, I'll structure the tool call with the name checkIpAddress and the ip\_address argument. No other tools are relevant here, like getStockLocations or checkInventory. Just need to output the tool call in the specified XML format. Let me double-check the syntax to avoid any errors. The JSON should have the name and arguments as specified. Alright, that should do it.
</think>
<tool\_call>
\{"name": "checkIpAddress",
    "arguments": "\{"ip\_address": "192.168.1.100"\}"\}
</tool\_call>

\vspace{2mm}  % 2毫米空行
tool: \{"status": "safe", "ip\_address": "192.168.1.100",   "reason": "Private IP address (RFC 1918 compliant)"\}

\vspace{2mm}  % 2毫米空行
assistant: 
<think>Okay, let me break this down. The user asked if the IP address 192.168.1.100 is flagged as a security risk. I called the checkIpAddress tool with that IP. The response says it's marked as safe because it's a private IP address under RFC 1918. Hmm, I need to explain this clearly. First, confirm the IP's status. Then, explain why it's safe. Maybe mention that private IPs are used internally and not routable on the public internet. Also, note that RFC 1918 defines these ranges. The user might not be familiar with RFC standards, so a brief explanation would help. Finally, offer further assistance if they need more checks. Keep it concise and friendly.</think>
The IP address **192.168.1.100** is **not flagged as a security risk** in our system. It is categorized as a **private IP address** (per [RFC 1918](https://tools.ietf.org/html/rfc1918)), which means it is reserved for internal networks and cannot be routed over the public internet. These addresses are inherently non-routable and considered safe by design. Would you like to check another IP address or investigate further?

\vspace{2mm}  % 2毫米空行
user: \#\#\#STOP\#\#\#

\end{promptbox}

\end{document}